\documentclass{wscpaperproc}
\usepackage{latexsym}
\usepackage{graphicx}
\usepackage{mathptmx}
\usepackage[T1]{fontenc}
\usepackage{tablefootnote}

\usepackage{amsmath}
\usepackage{amsfonts}
\usepackage{amssymb}
\usepackage{amsbsy}
\usepackage{amsthm}
\usepackage{mcite}

\usepackage[pdftex,colorlinks=true,urlcolor=blue,citecolor=black,anchorcolor=black,linkcolor=black]{hyperref}

\newtheoremstyle{wsc}
{3pt}
{3pt}
{}
{}
{\bf}
{}
{.5em}
{}

\theoremstyle{wsc}

\begin{document}

%
%
\WSCpagesetup{Gaur, Deshkar, Kshirsagar, Hayatnagarkar, and Venugopalan}

\title{Domain-driven Metrics for Reinforcement Learning: A Case Study on Epidemic Control using Agent-based Simulation}

\author{
\\
Rishabh Gaur\\
Gaurav Deshkar\\
Jayanta Kshirsagar\\
Harshal Hayatnagarkar\\
Janani Venugopalan\\ [12pt]
Engineering for Research (e4r)\\
Thoughtworks Technologies, Pune, INDIA\\
}

\maketitle
\begin{abstract}
For the development and optimization of agent-based models (ABMs) and rational agent-based models (RABMs), optimization algorithms such as reinforcement learning are extensively used. However, assessing the performance of RL-based ABMs and RABMS models is challenging due to the complexity and stochasticity of the modeled systems, and the lack of well-standardized metrics for comparing RL algorithms. In this study, we are developing domain-driven metrics for RL, while building on state-of-the-art metrics. We demonstrate our ``Domain-driven-RL-metrics'' using policy optimization on a rational ABM disease modeling case study to model masking behavior, vaccination, and lockdown in a pandemic. Our results show  the use of domain-driven rewards in conjunction with traditional and state-of-the-art metrics for a few different simulation scenarios such as the differential availability of masks.

\end{abstract}
\section{INTRODUCTION}
Agent Based-Models (ABMs) and Rational Agent-Based Models (RABMs) are being used extensively as modeling tools in various fields, including economics (~\shortciteNP{charpentier2021reinforcement},  ~\citeNP{arthur2021foundations}, ~\shortciteNP{cristelli2011critical}), social sciences (~\shortciteNP{bedson2021review}, ~\citeNP{peysakhovich2019reinforcement}, ~\citeNP{railsback2019agent}, ~\shortciteNP{hansen2019agent}), and behavior studies (~\citeNP{mabaso2021computationally}, ~\shortciteNP{kountouriotis2014agent}, ~\shortciteNP{taberna2020tracing}), due to their ability to simulate complex systems. The development and optimization of ABMs  and RABMs involve various modeling techniques and approaches, including rule-based models (~\shortciteNP{schluter2019potential}, ~\citeNP{macal2009agent}), reinforcement learning (RL) models (~\shortciteNP{charpentier2021reinforcement}, ~\shortciteNP{lee2017agent}, ~\shortciteNP{jalalimanesh2017simulation}), and network-based models ~\shortcite{carley2006biowar} and other AI models ~\shortcite{cervantes2020artificial}

RL is increasingly becoming the preferred method for developing RABMs and optimization of ABMs policy, owing to its ability to handle complex decision-making problems in dynamic environments. However, assessing the performance of RL-based ABMs and RABMS models is challenging due to the complexity and stochasticity of the modeled systems, and the lack of well-standardized metrics for comparing RL algorithms (~\shortciteNP{le2021metrics}, ~\shortciteNP{henderson2018deep}, ~\shortciteNP{chan2019measuring}). Unlike standard RL applications, RL-based ABMs and RABMs involve multiple agents interacting with each other and the environment, making it difficult to assess the performance of individual agents or the overall system ~\shortcite{SALTELLI201929}. Moreover, RL-based ABMs can generate multiple optimal policies or trajectories, further complicating the comparison of different algorithms. Thus, there is a need for standardized metrics and evaluation procedures that can provide a fair and consistent assessment of RL-based ABMs and RABMs.

In this study, we propose metrics to compare different RL algorithms used for the development and optimization of ABMs and RABMs. We illustrate our performance metrics using a case study on the public policy optimization of a rational agent-based epidemiological model for simulating sociological behavior during a COVID-19 epidemic. The remainder of the paper is structured as follows: Section \ref{related work} provides a review of the related literature, Section \ref{methods} describes the metrics for evaluating RL-based RABMs, followed by an epidemiology case study in Section \ref{case study}. We present the our case study on epidemic control using agent-based simulation and experimental results and finally, in Section \ref{conclusion}, we conclude the paper with a summary and discuss future work.

\section{RELATED WORKS}
The optimization of the outputs of ABMs and RABMs has gained traction in recent years, as the use of these models for decision-making has grown. As there are no standardized metrics in ABMs and RABMs while in Reinforcement Learning (RL) space, there are some common practices to evaluate and compare RL algorithms by using plots or tables of average cumulative reward (average returns), standard deviation of rewards, maximum mean-reward and recently, maximum reward over a fixed number of runs (episodes) ~\shortcite{henderson2018deep, oliveira2019difference, fernando2017pathnet, rusu2016progressive, parisotto2015actor, macua2017diff}. In RL space, the main problems stem from a knowledge gap of hyper-parameters, misleading and unstandardized evaluation metrics for the RL algorithms ~\shortcite{henderson2018deep, vamplew2011empirical}. \label{related work}

Therefore, researchers in RL space have explored different facets to unfold the black box nature of reinforcement learning and to compare and evaluate algorithms, e.g.: i) \textbf{Explainability}: Within this work, the authors analyze  the RL algorithm's interaction with the environment and understand the RL algorithm's underlying characteristics and aptitude in a task which make reinforcement learning more transparent, understandable, trustable and debuggable \cite{Sequeira2020} ii) \textbf{Reliability}: In this work, they proposed a set of metrics that measure reliability and aspects of the variability of RL algorithms.~\shortcite{chan2019measuring},  iii) \textbf{Algorithm completeness}: Here, they propose an algorithm evaluation metric of completeness, which states that an algorithm is complete on an environment, if the only required inputs to the algorithm is the meta-information about the environment( the number of state features and actions) and without hyperparameter tuning algorithm should reliably solve multiple tasks at different environments. ~\shortcite{jordan2020evaluating} etc. Some researchers also tried comparing RL algorithms on basis of data efficiency, goal-changing handling capability, trajectories, and policies \cite{atkeson1997comparison}. Other than the evaluation aspect, researchers are assessing existing reinforcement learning practices by numerous other factors i.e.: i) \textbf{Reproducibility}: For reproducibility, the authors focus on the difficulties faced by RL practitioners in reproducing a state-of-the-art deep RL method and similar results as most publications do not report all hyperparameters, proper in-depth implementation details, experimental setup, and evaluation methods for both baseline comparison methods and novel state-of-the-art work ~\shortcite{henderson2018deep}. ii) \textbf{Experimentation techniques}: Here, they are highlighting the issue of no standardized experimentation procedures in the RL community ~\shortcite{henderson2018deep}. iii) \textbf{Result reporting}: Here, researchers are pointing out one of the major issues in reinforcement learning literature which is the diversity of metrics, lack of significance testing, and statistical uncertainty leading to deceptive results reporting. (~\shortciteNP{agarwal2021deep}, ~\shortciteNP{henderson2018deep}, ~\shortciteNP{chan2019measuring}) iv) \textbf{Usability} of the algorithm across multiple environments including the time and effort spent in hyperparameter-tuning of an algorithm, and v) \textbf{Computational tractability} meaning that an  RL practitioner should be able to run the procedure and repeat experiments found in the literature ~\shortcite{jordan2020evaluating}.

In practice, RL algorithms are often evaluated and compared using one or the other metric based on reward, which is inadequate for a good comparison because of the volatile nature of the RL algorithm, environment stochasticity, and impact of noise on mean-reward (~\shortciteNP{henderson2018deep}, ~\shortciteNP{chan2019measuring}, ~\shortciteNP{islam2017reproducibility}). Reward alone may not provide a proper insight into an algorithm’s performance, it should be incorporated with domain knowledge to build more concrete evaluation metrics. Some researchers have tried comparing algorithms only using domain knowledge only ~\shortciteNP{nagendra2017comparison} but that is also inadequate for proper algorithm comparison. To the best of our knowledge, this is the first work towards evaluation metrics based on both domain knowledge and rewards. By considering both reward and domain knowledge, it adds more confidence and trust in the results. Our framework builds on this prior work by providing a set of metrics for analyzing the different RL algorithms used in RABMs and for optimizing ABMs outputs.

\section{METRICS TO EVALUATE RL-BASED RABMs AND ABMs OPTIMIZATION}
\begin{figure}[ht]
  \centering
  \includegraphics[width=\textwidth]{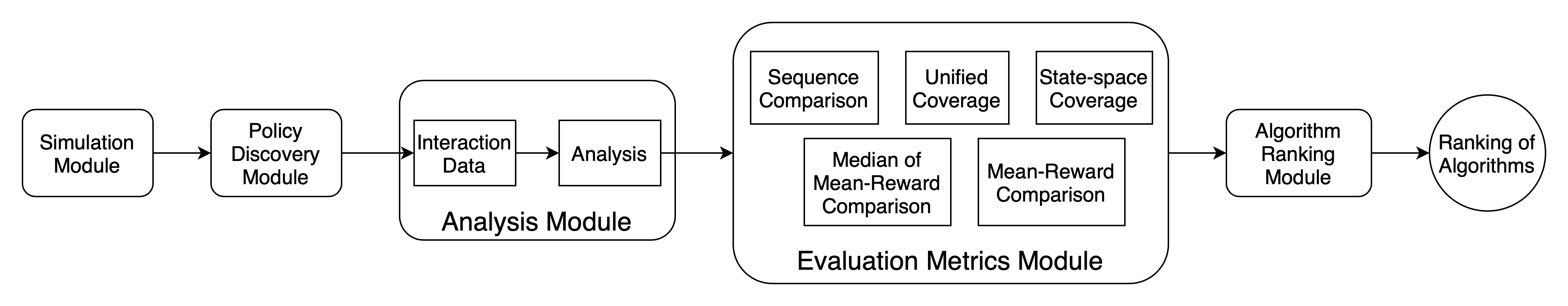}
  \caption{RL metric calculation from agent-based epidemic simulator with the 9-Compartment model, individual decision-making components, and public policy optimization.}
  \label{block_diagram}
\end{figure}

Figure \ref{block_diagram} shows the workflow of our algorithm ranking, it starts with the ``simulation module'' where we perform agent-based simulations for a small community of 1000 rational individuals. The ``policy discovery module'' is responsible for generating optimal policy to control the pandemic situation. Next is the ``Analysis module'', which is developed using InterestingnessXRL (a Python library for eXplainable Reinforcement Learning (XRL) \cite{Sequeira2020}), which extracts interaction data (data about the interaction of an RL algorithm with its environment) and performs analysis on that data to generate indicators of an RL algorithm's performance. Then, those indicators are used to generate our evaluation metrics. ``Evaluation metric module'' provides the algorithm rankings based on each metric. At the ``algorithm ranking module'', the aggregate composite rank of every algorithm is calculated for the final ranking. \label{methods}

\subsection{Analysis for developing RL metrics}
Our study expands on the work by Sequeira \emph{et. al.} \cite{Sequeira2020} to extract interaction data and analyze the RL algorithms' behavior and performance during training. The evaluation metrics are based on the following analyses:

\textbf{State Frequency Analysis} provides an analysis of an RL algorithm’s state visiting pattern. It extracts coverage of state space, dispersion of visited states, and frequent/infrequent states. 

\textbf{State-Action Frequency Analysis} provides analysis of an RL algorithm’s history of action selection/execution with the environment and extracts coverage of state-action space, and mean dispersion of the execution of actions in visited states and certain/uncertain states. 

\textbf{Reward Analysis} extracts state-action pairs that are, on average among all pairs, significantly more or less rewarding than others. It provides us with those situations where the RL algorithms performed very well or did not perform well enough to receive an average reward.

\textbf{Transition Value Analysis} represents an analysis of an RL algorithm's state-value function about state transitions. The goal is to analyze how the value attributed to some observation changes about possible observations taken at the next time step. It calculates the local and absolute extrema states, i.e., the states whose values are greater/lower than all possible next states, respectively.

\textbf{Sequence Analysis} calculates the common and important sequences of actions learned by the RL algorithm during interactions with its environment. In particular, interesting sequences involve starting from a local minima state, then executing the most likely action, and then performing actions to reach a local maximum (target) state. Only target states that are reachable with a minimum probability are considered. The most valuable target state is chosen as the one with the highest product between the probability and state Q value. The exploit run’s sequence represents how well an RL algorithm has learned to solve the given reinforcement learning problem in a given environment. On basis of domain knowledge, the end state (local maxima state) of sequences tells us how well the RL algorithm has performed. 
    
\subsection{Domain-driven Metrics}
The domain-driven metrics are as follows:

\textbf{Sequence Comparison} is a domain knowledge-based metric, as it compares algorithms based on the percentage of the best sequences across exploit runs (test runs). A sequence is considered the best sequence when it ends with the best end state, which is decided on using domain knowledge. An algorithm with a higher best sequence percentage will be highly ranked.

\textbf{Median of Mean-Rewards} compares algorithms based on the median of mean-rewards across exploit runs (test runs) and provides an algorithm ranking. An algorithm with a higher median reward value will have a better rank than other algorithms.

\textbf{State-space Coverage}
State-space coverage provides the percentage of the total states/state indices (unique combinations of binned state-space components decided on basis of domain-knowledge) that an RL algorithm has visited during training and ranks algorithms on basis of that coverage percentage. The higher the coverage during training, the better the algorithm’s performance.

\textbf{Unified Coverage}
Unified coverage provides the algorithm ranking based on the unified coverage percentage, which is the unified coverage of both state and state-action space during the training. The higher the coverage during training, the better the algorithm’s performance.

\textbf{Mean-Reward Comparison}
Mean-reward comparison metric provides us with the algorithm ranking based on the mean rewards received by the RL algorithm during the training period. Higher the mean reward during training better the performance.\\

To get the final performance ranking of algorithms for an experiment, the aggregate rank of every algorithm is calculated using the ranks based on all five metrics. Then, based on the aggregate rank final performance ranking is calculated.

\section{CASE STUDY: COMPARING RL ALGORITHMS FOR OPTIMIZATION OF RABMs}
\label{case study}
\subsection{Simulation Model}
We demonstrate the utility of our metrics by using an agent-based simulation for a small community of 1000 rational individuals belonging to different age groups (0-17, 18-59, 60-99) as a case study. The individuals in our simulation can make their own decisions about going to the office/school (depending on their age), shopping, wearing masks, using public transportation, and staying at home.  Agents older than 30 are considered employed, and those younger than 30 are students. Every agent follows a schedule. A schedule is defined  for 24 hours, with six vector ticks in the simulation. Agents spend  8 hours (2 vector ticks) at home, followed by 12 hours (3 vector  ticks) at either the office or school based on their age, followed  by 4 hours (1 vector tick) shopping or at home based on their  preference. We chose these locations to demonstrate a minimalist  model where people are moving to meet the same people routinely (offices, houses) and different people intermittently (shops). Through this simulation, we are trying to simulate the spread of COVID in a small community of 1000 individuals, where policymakers can make interventions of lockdown and vaccination drives, by observing the infections, hospitalizations, and the economy to control the pandemic situation.

\subsection{Simulation Experiments}
To look at the efficacy of the RL comparison metrics across experiments, we performed simulations with slightly different epidemiological scenarios. Using the above minimalist epidemic model we executed a simulation with 1000 agents for 100 days, i.e., 600 simulation ticks. The policy is updated every seven days, i.e., every 42 simulation ticks. Our experiments have two types of vaccines with 80\% and 60\% effectiveness and two types of masks with 80\% and 40\% effectiveness. We have performed three experiments on top of our epidemic model.

\begin{table}[htbp]
    \caption{Vaccines and Masks Availability For Experiments}
    \begin{center}
    \begin{tabular}{|c|c|c|c|c|}
    \hline
    \textbf{Name} & \textbf{Vaccine 1 }& \textbf{Vaccine 2}  & \textbf{Mask 1} & \textbf{Mask 2} \\
    \hline
     Baseline & 6 & 6 & 500 &  1000  \\
     High Mask & 6 & 6 & 800 & 1000  \\
     Low Mask & 6 & 6 & 100 & 1000  \\
    \hline
 \multicolumn{5}{l}{\textbf{*Vaccines doses available per day}}
    \end{tabular}
    \label{tab:vac-mask-availability}
    \end{center}
\end{table}

\textbf{Baseline Experiment}
Every individual can choose to get vaccinated on any day. Also, every individual can have a low-efficiency mask (1000 masks) but only 50\% of the population (500 masks) can have a high-efficiency mask.

\textbf{High-Mask Experiment}
We increased high-efficiency mask availability to 800 masks which means now 80\% of the population can have high-efficiency masks and the rest is the same as in the baseline experiment.

\textbf{Low-Mask Experiment}
We decreased high-efficiency mask availability to 100 masks which means now only 10\% of the population can have high-efficiency masks. 

\subsection{Policy Optimization to Control the Epidemic}
In this study, we use RL algorithms to optimize the total infections, hospitalizations, and economic status of the population. To derive, the optimal policy in each simulation scenario, we used RL algorithms. In our study, we consider the policymaker and the optimization of public policy as a Markov Decision Process (MDP) with a continuous action space to account for factors such as lockdown start and end days, vaccination start and end days, and mask availability. A continuous space-action MDP is defined as ($S, A, P, R, \gamma $), where $s \in S$ is a finite state space, $a \in A$ is a finite action space, $P \equiv P(s_{t+1}|s_{t}, a_{t})$ is a transition kernel which is continuous in $a$. $R \equiv r(s_{t}, a_{t})$ is a reward function continuous in $a$, and $\gamma \in (0, 1)$ is the discount factor. While in a traditional MDP, the actions ($a \in A$) and the rewards ($r(s_{t}, a_{t})$) are considered certain, in our study, we consider that while the directives of public policy (actions), such as the exact lockdown start, and vaccination, cannot be followed, the measurement of states, such as the number of infected, is also not precise. As a result, we introduce uncertainty in the actions and the state spaces. We tested these variants for deep deterministic policy gradient (DDPG) and Twin Delayed DDPG (TD3) algorithms. This gave us eight RL algorithms to compare, four versions of each DDPG and TD3 namely Vanilla (DDPG/ TD3), Uncertainty in Actions (NR\_DDPG/NR\_TD3), Uncertainty in States (BN\_DDPG/ BN\_TD3) and Uncertainty in Actions and States (NR\_BN\_DDPG/ NR\_BN\_TD3) ~\shortcite{kdd}. As mentioned above, the major challenge for this is to pick the best-performing algorithm and to get their performance ranking for a given environment (experiment). 

\subsection{Comparison of RL Techniques using Domain Driven Metrics}
 We are showing results for High and Low Mask experiments in Table \ref{tab:high-mask-algo-ranking} and \ref{tab:low-mask-algo-ranking} respectively.
 
\subsubsection{Mean Rewards}
\begin{figure}[ht]
  \centering
  \includegraphics[width=0.8\linewidth]{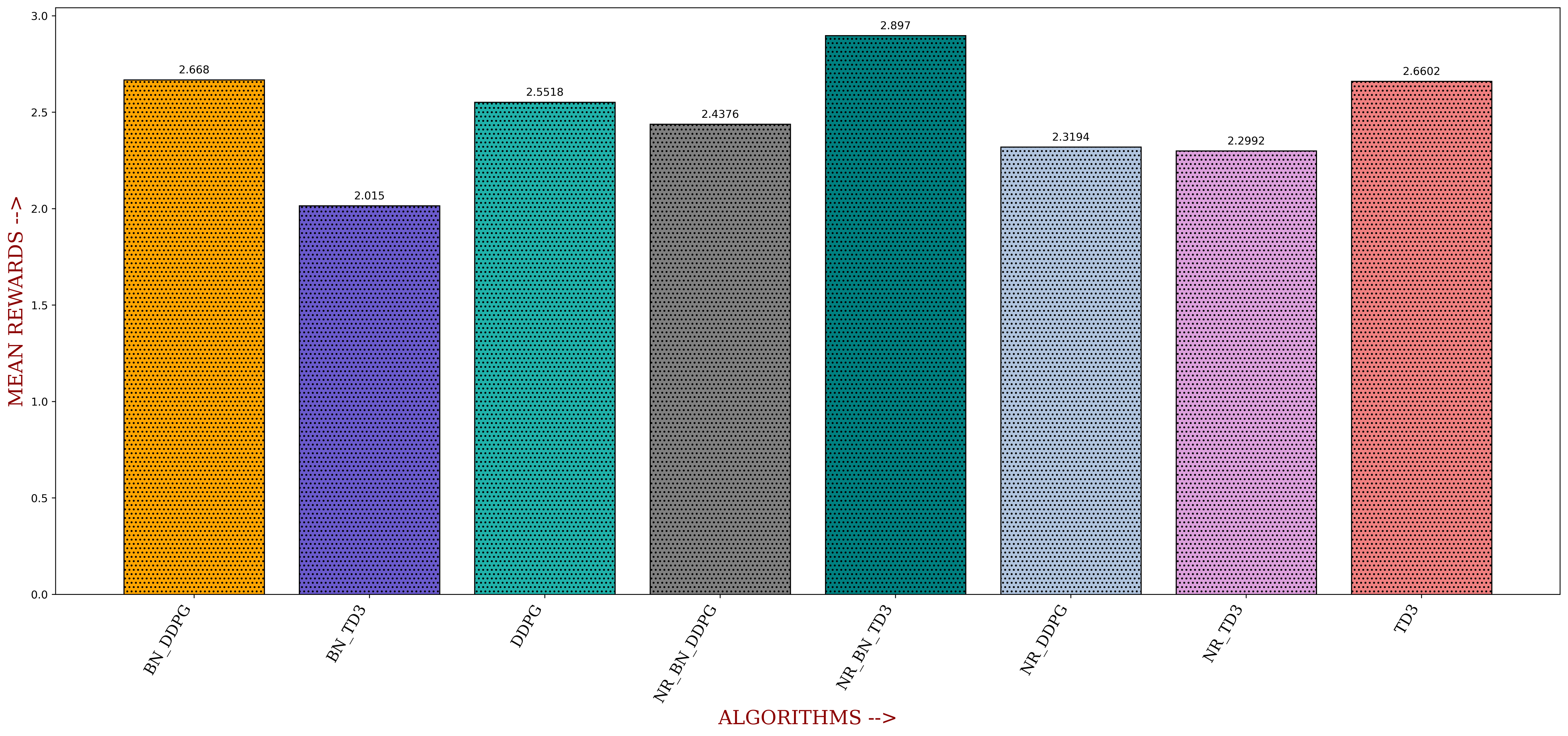}
  \caption{Mean of mean-rewards across experiments for algorithms}
  \label{mean_rwd_plot}
\end{figure}

Mostly in the RL space, the mean reward is used as the primary metric to evaluate and compare RL algorithms. If we’re comparing RL algorithms only on the basis of their mean rewards (across training), we might not get the actually best-performing algorithm. Algorithms may have very similar mean rewards across training (Figure \ref{mean_rwd_plot}) or the best algorithm according to the mean reward, might not have explored the environment well. Alternatively, the actions may be highly rewarding in the short term but not in the long term. To mitigate the effects of this phenomenon,  we compare and rank RL algorithms using a compositive metric of five different metrics namely state space coverage, unified coverage, exploit run’s sequence and median-reward comparison, and training-level mean-reward comparison.

While the explainability study by Sequeira \emph{et. al.} is compatible with discrete space only, we extend our work to continuous spaces and use the analysis to develop metrics. The  first step in this is to perform binning on states and actions. 

\subsubsection{State-space} \label{state_space}
The state consists of three state-space components which are normalized to (0, 1). The first component is \textbf{InfectedMild} which is the number of individuals with mild infection of the disease. Second, \textbf{Hospitalized} represents the number of individuals hospitalized because of the epidemic and the last is \textbf{minStockHousePercentage} which is the number of families with house stock below-defined threshold, which represents a significant percentage of families can fall below the poverty line (BPL). Now, to convert these three continuous state-space components to discrete, we performed binning for each component and then assign unique values to all combinations of indices of the bins, and those unique values are called state indices. Bins used to perform binning of state components are \textbf{[0.0, 0.05, 0.10, 0.15, 0.20, 1]} which is decided on the basis of domain knowledge i.e.: that there will be a small number of situations when more than 20\% population is below the poverty line or hospitalized. So, we have not considered finer bins for the range (0.20, 1). 

\textbf{Example} :
State Index 0 represents \textbf {[0.0 - 0.05, 0.0 - 0.05, 0.0 - 0.05]} unique combination of state components which means:

\begin{enumerate}
    \item{\textbf{0.0 - 0.05}: 0-5\% of the population is infected with Mild Symptoms.}
    \item{\textbf{0.0 - 0.05}: 0-5\% of the population is hospitalized because of COVID.}
    \item{\textbf{0.0 - 0.05}: 0-5\% of the houses are suffering from a severe financial crisis, they are Below the Poverty Line (BPL) as their house stocks are below defined thresholds.}
    \label{state-0}
\end{enumerate}

\subsubsection{Action-space} \label{action_space}
The action consists of eight continuous action-space components which are in the range of 0-7 days as  the policy is updated every week (42 simulation ticks). The initial two are lockdown start date and duration, then the vaccination drive start date and duration for age different age groups (0-17, 18-59, 60-99). We need to convert these eight continuous action-space components to discrete, for which we follow a similar process that we perform for state-space components and generate action indices, which represent unique combinations of bin indices of action-space components. Bins used to perform binning of action-space components are \textbf{[0, 2.5, 5, 7]} which are also decided on basis of some domain-knowledge.

\textbf{Example} :
Action Index 2432 represents \textbf{[2.5 - 5, 0 - 2.5, 2.5 - 5, 0 - 2.5, 0 - 2.5,   0 - 2.5, 0 - 2.5, 5 - 7]} unique combination of action-space components which means:

\begin{enumerate}
    \item{\textbf{2.5 - 5}: Lockdown should be started between the second half of 2\textsuperscript{nd} day to 5\textsuperscript{th} day of a week.}
    \item{\textbf{0 - 2.5}: Lockdown should be imposed for a duration of 0 to 2.5 days.}
    
    \item{\textbf{2.5 - 5}: Vaccination drives for the age group of 0-17 yrs should be started between the second half of 2\textsuperscript{nd} day to 5\textsuperscript{th} day of a week.}
    \item{\textbf{0 - 2.5}: Duration of vaccination drives for the age group of 0-17 yrs should be 0 to 2.5 days.}
    
    \item{\textbf{0 - 2.5}: Vaccination drives for the age group of 18-59 yrs should be started before the second half of the 2\textsuperscript{nd} day of a week.}
    \item{\textbf{0 - 2.5}: Duration of vaccination drives for the age group of 18-59 yrs should be 0 to 2.5 days.}
    
    \item{\textbf{0 - 2.5}: Vaccination drives for the age group of 60-99 yrs should be started before the second half of the 2\textsuperscript{nd} day of a week.}
    \item{\textbf{5 - 7}: Duration of vaccination drives for the age group of 60-99 yrs should be 5 to 7 days.}
    \label{action-2432}
\end{enumerate}

\begin{table}[htbp]
    \caption{High Mask Experiment: Domain-driven metrics based algorithm ranking}
    \scalebox{0.75}{
    \begin{tabular}{|c|c|c|c|c|c|c|c|}
    \hline
     \textbf{Algorithm} & \textbf{Mean Reward} &  \textbf{State Coverage} & \textbf{Unified Coverage} & \textbf{Best sequences \%} & \textbf{Median Reward} &  \textbf{Aggregate Rank} &  \textbf{Rank} \\
    \hline
           TD3 &        2.853 &               28.889 &            \textbf{68.744} &                      34.55 &        2.8135 &    \textbf{9} &   \textbf{1.0} \\
     NR\_BN\_TD3 &  \textbf{2.940} &           11.111 &                   46.280 &                   \textbf{100.00} &          \textbf{2.9085} &              14 &   2.0 \\
          DDPG &        2.938 &                11.111 &                   45.960 &                   \textbf{100.00} &                 2.8995 &              15 &   3.0 \\
       BN\_DDPG &        2.871 &               15.556 &                   54.152 &                      78.57 &                 2.8115 &              15 &   3.0 \\
    NR\_BN\_DDPG &        1.946 &       \textbf{31.111} &                   48.608 &                       0.00 &                 1.9340 &              18 &   4.0 \\
        BN\_TD3 &        1.920 &               28.889 &                   58.240 &                       0.00 &                 1.8845 &              21 &   5.0 \\
       NR\_DDPG &        1.926 &               24.444 &                   54.384 &                       0.00 &                 1.9065 &              22 &   6.0 \\
        NR\_TD3 &        1.927 &               15.556 &                   53.984 &                       0.00 &                 1.8965 &              26 &   7.0 \\
    \hline
    \end{tabular}}
    \label{tab:high-mask-algo-ranking}
\end{table}

\begin{table}[htbp]
    \caption{Low Mask Experiment: Domain-driven metrics based algorithm ranking}
    \scalebox{0.75}{
    \begin{tabular}{|c|c|c|c|c|c|c|c|}
    \hline
     \textbf{Algorithm} & \textbf{Mean Reward} &  \textbf{State Coverage} & \textbf{Unified Coverage} & \textbf{Best sequences \%} & \textbf{Median Reward} &  \textbf{Aggregate Rank} &  \textbf{Rank} \\
    \hline
       NR\_DDPG &        2.894 &         \textbf{51.111} &           \textbf{59.064} &                      63.89 &         \textbf{2.8730} &    \textbf{5} & \textbf{1.0} \\
        NR\_TD3 & \textbf{2.900} &           15.556 &                   56.336 &                \textbf{80.00} &                 2.8650 &              11 &   2.0 \\
     NR\_BN\_TD3 &        2.898 &           20.000 &                   57.528 &                      76.00 &                 2.8595 &              12 &   3.0 \\
          DDPG &        2.868 &             44.444 &                   55.520 &                      53.19 &                 2.8595 &              15 &   4.0 \\
    NR\_BN\_DDPG &        2.039 &           42.222 &                   58.368 &                       0.00 &                 2.1490 &              20 &   5.0 \\
           TD3 &        2.878 &             17.778 &                   55.040 &                      71.88 &                 2.8150 &              22 &   6.0 \\
        BN\_TD3 &        1.904 &            24.444 &                   55.352 &                       0.00 &                 1.8520 &              27 &   7.0 \\
       BN\_DDPG &        1.900 &            31.111 &                   55.328 &                       0.00 &                 1.8520 &              28 &   8.0 \\
    \hline
    \end{tabular}}
    \label{tab:low-mask-algo-ranking}
\end{table}

\subsection{Domain-driven Metrics}

Results of algorithm ranking based on Domain-driven metrics for High and Low mask experiments are shown in Table \ref{tab:high-mask-algo-ranking} and \ref{tab:low-mask-algo-ranking} respectively, where it exhibits the value of mean and median rewards, state and unified coverage percentages and best sequence percentage for algorithms with their aggregate rank across all metrics and the final ranking.

\subsubsection{Sequence Comparison}
Sequence comparison is a domain knowledge-based metric, algorithms are compared based on the best sequence percentage. The best sequence is a sequence ending with the best end state, which is decided on the basis of domain knowledge. In our experiment, state index 0 is the best end state because at this state infected, hospitalized, and below the poverty line (BPL) population is the least (< 5\%). The sequence is a chain of state, action, next state, next action, and so on, where the starting state is the local minima state and the ending state is the local maxima state (extracted during transition value analysis).
\\
\\
\textbf{Example of a sequence} :

Sequence: \textbf{[Run-109-Exploit, 100, 2432, 50, 2435, 25, 2431, 0]}

\begin{enumerate}
    \item{\textbf{Run-109-Exploit}: Name of the run, which is 109th run in which exploitation is done.}
    
    \item{\textbf{100}: Local minima state index (start state) for this run which represents \textbf{[0.2 - 1.0, 0.0 - 0.05, 0.0 - 0.05]} state-components combination.}
    
    \item{\textbf{2432}: Action taken by an RL algorithm at the start state, which represents \textbf{[2.5 - 5, 0 - 2.5, 2.5 - 5, 0 - 2.5, 0 - 2.5, 0 - 2.5, 0 - 2.5, 5 - 7]} action-space components combination.}
    
    \item{\textbf{50}: Next state index visited by an RL algorithm on taking action index 2432 at start-state. This state index represents \textbf{[0.1 - 0.15, 0.0 - 0.05, 0.0 - 0.05]} state-components combination.}
    
    \item{\textbf{2435}: Action taken by an RL algorithm at state index (50), which represents \textbf{[2.5 - 5, 0 - 2.5, 2.5 - 5, 0 - 2.5, 0 - 2.5, 0 - 2.5, 2.5 - 5, 5 - 7]} action-space components combination.}
    
    \item{\textbf{25}: Next state index visited by an RL algorithm. This state index represents \textbf{[0.05 - 0.1, 0.0 - 0.05, 0.0 - 0.05]} state-components combination.}
    
    \item{\textbf{2431}: Action taken by an RL algorithm at state index (25), which represents \textbf{[2.5 - 5, 0 - 2.5, 2.5 - 5, 0 - 2.5, 0 - 2.5, 0 - 2.5, 0 - 2.5, 2.5 - 5]} action-space components combination.}
    
    \item{\textbf{0}: End state of the sequence which is the local maxima state for this run. This state index represents \textbf{[0.0 - 0.05, 0.0 - 0.05, 0.0 - 0.05]} state-components combination.}
\end{enumerate}

For high mask experiment (Table \ref{tab:high-mask-algo-ranking}), NR\_BN\_TD3 and DDPG have the best sequence percentage of 100\% which means all exploit run sequences are ending with the best end state index (which is state index \hyperref[state-0]{0} in our case) which tells us that these algorithms are able to control the epidemic situation well as the sequence is ending with state 0 which is the best state according to domain knowledge as infected, hospitalized and BPL population is minimum at this state. For low mask experiment (Table \ref{tab:low-mask-algo-ranking}), NR\_TD3 has the highest best sequence percentage of 80\%.   

\subsubsection{Median of Mean-Rewards Comparison}
In this metric, algorithms are compared and ranked on basis of the median of mean rewards across exploit runs. For the high mask experiment (Table \ref{tab:high-mask-algo-ranking}), NR\_BN\_TD3 has received the highest median reward of 2.9085 followed by DDPG and TD3 with 2.8995 and 2.8135 respectively. For low mask experiment (Table \ref{tab:low-mask-algo-ranking}), NR\_DDPG has received the highest median reward of 2.8730 followed by NR\_TD3, DDPG, NR\_BN\_TD3 with slightly lower median-reward of 2.8650, 2.8595 and 2.8595 respectively. 

\subsubsection{State-space Coverage}
This metric compares and ranks algorithms based on the state-space coverage during training. As discussed in Section \ref{state_space}, that state has 3 components that need to be binned using 5 bins ([0.0, 0.05, 0.10, 0.15, 0.20, 1]), which give 5\textsuperscript{3} = 125 unique combinations of bin indices. Based on the domain knowledge, only 45 combinations (state indices) out of 125 combinations are valid, which means the RL algorithm will never be able to visit those 80 combinations and that is impacting the state-coverage percentage. Only 45 state indices are valid because at max 10\% population can be hospitalized (environment property). According to WHO \cite{WHO}, the highest hospital bed-to-population ratio is 143:10000 which is 1.43\% only, so we also incorporate real-world scenarios in our simulation and revise the state coverage to reflect this. For high mask experiment (Table \ref{tab:high-mask-algo-ranking}), NR\_BN\_DDPG has the highest coverage of 28.889\%, which means ~29\% of state-space component's bin combinations (state indices) are explored by an RL algorithm during training. For the low mask experiment (Table \ref{tab:low-mask-algo-ranking}), NR\_DDPG has covered the highest number of state components bin combinations which is 51.11\%.

\subsubsection{Unified Coverage}
Unified coverage is the combination of State and State-Action space coverage. For the high mask experiment (Table \ref{tab:high-mask-algo-ranking}), vanilla TD3 outperformed all other algorithms in unified coverage percentage. Vanilla TD3 got  68.744\% of unified coverage percentage, which signifies that TD3 has maximum exploration of State and  State-Action space, followed by BN\_TD3, NR\_DDPG, BN\_DDPG, and NR\_TD3 with 58.240, 54.384, 54.152 and 53.984 respectively. For low mask experiment (Table \ref{tab:low-mask-algo-ranking}), NR\_DDPG has the highest unified coverage percentage of 59.064\% followed by NR\_BN\_DDPG, and NR\_BN\_TD3 with 58.368\%, and 57.528\% respectively.

\subsubsection{Mean-Reward Comparison}
This metric compares algorithms based on their mean rewards during training. For the high mask experiment (Table \ref{tab:high-mask-algo-ranking}), NR\_BN\_TD3 has received the highest mean-reward of 2.940 followed by vanilla DDPG with a very slight difference of 0.002 having a mean reward of 2.938. Such a situation raises questions about the robustness of the existing evaluation metric in reinforcement learning space because if we are ranking algorithms only based on mean rewards during training and if the DDPG algorithm took a few better actions then it might be the case that the ranking order got flipped. Similarly, for low mask experiment (Table \ref{tab:low-mask-algo-ranking}), NR\_TD3, NR\_BN\_TD3, and NR\_DDPG have their mean-rewards in very close proximity as it is 2.900, 2.898 and 2.894 mean-rewards respectively for these three algorithms.

\begin{table}[htbp]
    \caption{High Mask Exp.: Domain-driven and Google's Reliability Metrics based algorithm ranking}
    \scalebox{0.65}{
    \begin{tabular}{|c|c|c|c|c|c|c|c|c|}
    \hline
     \textbf{Algorithm} & \textbf{IQR} &  \textbf{LCVaRonDiff} & \textbf{LCVaRonDrawDown} & \textbf{Median Performance} & \textbf{Reliability Rank} & \textbf{Domain Rank} & \textbf{Aggregate Rank} &  \textbf{Rank} \\
    \hline
        TD3 &  1.0 &              8.0 &                  1.0 &                 4.0 &              14.0 &                   9 &            23.0 &   1.0 \\
        NR\_BN\_TD3 &  4.0 &              5.0 &                  4.0 &                 2.0 &              15.0 &                  14 &            29.0 &   2.0 \\
        DDPG &  6.0 &              3.0 &                  6.0 &                 2.0 &              17.0 &                  15 &            32.0 &   3.0 \\
        BN\_DDPG &  8.0 &              1.0 &                  8.0 &                 2.0 &              19.0 &                  15 &            34.0 &   4.0 \\
        NR\_BN\_DDPG &  5.0 &              4.0 &                  5.0 &                 6.5 &              20.5 &                  18 &            38.5 &   5.0 \\
        NR\_DDPG &  3.0 &              6.0 &                  3.0 &                 6.5 &              18.5 &                  22 &            40.5 &   6.0 \\
        NR\_TD3 &  2.0 &              7.0 &                  2.0 &                 6.5 &              17.5 &                  26 &            43.5 &   7.0 \\
        BN\_TD3 &  7.0 &              2.0 &                  7.0 &                 6.5 &               22.5 &                  21 &            43.5 &   7.0 \\
    \hline
    \end{tabular}}
    \label{tab:reliability-high-mask-algo-ranking}
\end{table}

\begin{table}[htbp]
    \caption{Low Mask Exp.: Domain-driven and Google's Reliability Metrics based algorithm ranking}
    \scalebox{0.65}{
    \begin{tabular}{|c|c|c|c|c|c|c|c|c|}
    \hline
     \textbf{Algorithm} & \textbf{IQR} &  \textbf{LCVaRonDiff} & \textbf{LCVaRonDrawDown} & \textbf{Median Performance} & \textbf{Reliability Rank} & \textbf{Domain Rank} & \textbf{Aggregate Rank} &  \textbf{Rank} \\
    \hline
        NR\_DDPG &  4.0 &              5.0 &                  4.0 &                 3.0 &              16.0 &                   5 &            21.0 &   1.0 \\
        NR\_TD3 &  3.0 &              6.0 &                  3.0 &                 3.0 &              15.0 &                  11 &            26.0 &   2.0 \\
        NR\_BN\_TD3 &  5.0 &              4.0 &                  5.0 &                 3.0 &              17.0 &                  12 &            29.0 &   3.0 \\
        DDPG &  7.0 &              2.0 &                  7.0 &                 3.0 &              19.0 &                  15 &            34.0 &   4.0 \\
        TD3 &  2.0 &              7.0 &                  2.0 &                 3.0 &              14.0 &                  22 &            36.0 &   5.0 \\
        NR\_BN\_DDPG &  6.0 &              3.0 &                  6.0 &                 7.0 &              22.0 &                  20 &            42.0 &   6.0 \\
        BN\_DDPG &  1.0 &              8.0 &                  1.0 &                 7.0 &              17.0 &                  28 &            45.0 &   7.0 \\
        BN\_TD3 &  8.0 &              1.0 &                  8.0 &                 7.0 &              24.0 &                  27 &            51.0 &   8.0 \\
    \hline
    \end{tabular}}
    \label{tab:reliability-low-mask-algo-ranking}
\end{table}
\footnotesize{\textbf{NOTE:- LCVaR:} Lower Conditional Value at Risk, \textbf{IQR:} Inter Quartile Range}

\subsection{Algorithm Ranking}
Vanilla TD3 has outperformed all other algorithms with a large aggregate rank difference for the high mask experiment (Table \ref{tab:high-mask-algo-ranking}). Vanilla TD3 has performed consistently across all 5 evaluation metrics to achieve the lowest aggregate rank of 9 followed by NR\_BN\_TD3, DDPG, and BN\_DDPG with the aggregate rank of 14, 15, and 15 respectively. For low mask experiment (Table \ref{tab:low-mask-algo-ranking}), NR\_DDPG is ranked 1 with a  margin of 6 aggregate rank, followed by NR\_TD3, and NR\_BN\_TD3 with the aggregate rank of 11 and 12 respectively.

\subsection{Comparison with State of the Art Metrics}
After we compute our Domain-driven metrics, we compared our work with other RL metrics by Chan \emph{et. al.} ~\shortcite{chan2019measuring}. Then, we additionally incorporated the metrics by Chan \emph{et. al.} into our ranking system. Chan \emph{et. al.} measure dispersion and risk using interquartile range (IQR) and conditional value at risk (CVaR) scores.\\

\textbf{Dispersion} is computed using the IQR metric. For the IQR metric, the rewards are detrended prior to IQR calculations. The IQR within a sliding window was chosen to keep the metric agnostic of distributions, and the detrending (i.e., $y_{t}' = y_{t} - y_{t-1})$) was done to preserve the positive increase in rewards while training. This metric was computed a few times (early, middle, and end) during training---the lower this value, the better the model.\\

 \textbf{Risk}: The conditional value at risk (CVaR)  evaluates the expected loss value in worst-case scenarios, parameterized by a quantile $\alpha_{q}$. This gives the risk as the heaviness of the lower tail of the distribution. We compute three risk scores, overall risk over time, Short-term Risk across Time (SRT), and Long-term Risk across Time (LRT) ~\shortcite{chan2019measuring}. SRT is computed using CVaR on Differences and allows us to measure the most extreme short-term drop over time. LRT is computed using CVaR on Drawdown and allows us to measure the potential of the algorithm to lose a lot of performance relative to its peak over time.

Then a consolidated reliability score is computed as a combination of the dispersion and the risk scores. Google's RL Reliability metrics-based algorithm ranking is combined with our Domain-driven metrics to calculate overall aggregate rank, on basis of which algorithms are ranked. Algorithm rankings based on Domain-driven metrics and Reliability metrics incorporated with Domain-driven metrics are very much alike, especially the rank of best-performing algorithms is exactly the same which adds more confidence in the result. Results of algorithm ranking based on Reliability metrics incorporated with Domain-driven metrics for High and Low mask experiments are shown in Table \ref{tab:reliability-high-mask-algo-ranking} and \ref{tab:reliability-low-mask-algo-ranking} with ranks according to all 4 Reliability metrics (i.e.: IQR, LCVaRonDiff, LCVaRonDrawDown, and Median Performance) and the aggregate rank according to  Reliability and Domain-driven metrics, followed by the overall aggregate rank and final ranks. For the high mask experiment (Table \ref{tab:reliability-high-mask-algo-ranking}), the ranking of TD3, NR\_BN\_TD3, DDPG, NR\_TD3, and NR\_DDPG is exactly the same as before, the rank of BN\_DDPG, and NR\_BN\_DDPG got shifted down by one and BN\_TD3 rank decreased from 5 to 7. For Low mask experiment (Table \ref{tab:reliability-low-mask-algo-ranking}), the rank of NR\_DDPG, NR\_TD3, NR\_BN\_TD3, and DDPG remains the same as before but the rank of BN\_DDPG, and TD3 got shifted up by one and NR\_BN\_DDPG, and BN\_TD3 rank decreased by one. 

In Figure \ref{mean_rwd_plot}, demonstrates that the mean of mean-rewards cannot be used to compare the different RL algorithms due to low range and the environment stochasticity. The algorithm ranking showed a larger variability with the addition of our Domain-driven metrics. This shows that while the mean-rewards-based ranking is not robust, the algorithm ranking based on our Domain-driven metrics is relatively robust as it stays similar to the ranking when the state-of-the-art metrics are incorporated with Domain-driven metrics. This suggests a novel approach to RL evaluation metrics that combines reward and domain knowledge can explore additional facets of an RL algorithm training and testing which any single metric does not independently achieve. In addition,  such a composite metric can also provide a source of trust to the ultimate users of the ABM the policymakers who derive policy using ABMs.

\section{CONCLUSION}
The discovery of an optimal RL algorithm for epidemiological simulation policy is particularly challenging due to the lack of well-established metrics. Will the increasing use of RL in epidemiology, the traditional reward-based metrics are not effective for comparing the  different algorithms. This work proposes domain knowledge-driven metrics for RL method comparison, which can be used in the domains such as epidemiology. Our case study illustrates public health and economy optimization in the presence of rational individual choices and uncertainty in state and action. We compared eight different algorithm choices to pick the performing algorithm. We demonstrated our results using a single model in a small community. Shortly, we will expand the model to include a larger scale of rational agents with more complex human and economic behavior. In addition, we do not employ explainability or techniques to explain each model's decision. Shortly, we aim to extend our work to account for explainability.\label{conclusion}

\bibliographystyle{wsc}

\bibliography{references}

\section*{AUTHOR BIOGRAPHIES}

\noindent {\bf RISHABH GAUR} is a Data Scientist at Engineering for Research (e4r), Thoughtworks, India. He holds a Bachelor of Technology with Specialization in Big Data \& Analysis (IBM) degree from Dehradun Institute of Technology, with more than two years of experience in data science, NLP, and RL. His email address is \email{
rishabh.gaur@thoughtworks.com}.\\

\noindent {\bf GAURAV DESHKAR} is a Senior Application Developer at e4r, Thoughtworks, India. He holds a bachelor’s in Computer Applications, with 6+ years of experience in software development. His work is on retail, healthcare, and simulation. His email is \email{gauravd@thoughtworks.com}.\\

\noindent {\bf JAYANTA KSHIRSAGAR} is a research engineer at e4r, Thoughtworks, India. He is passionate about understanding complex systems using agent-based simulation models. His email is \email{jayantak@thoughtworks.com}.\\

\noindent {\bf HARSHAL HAYATNAGARKAR} is a Principal Computer Scientist and the Head Scientist at e4r, Thoughtworks, India. He has diverse research interests such as complex systems, accelerating scientific and policy discovery, simulating large-scale social dynamics, and modeling human behavior. His email is \email{harshalh@thoughtworks.com}.\\

\noindent {\bf JANANI VENUGOPALAN} is a Lead Data Scientist at Engineering for Research (e4r), Thoughtworks, India. She holds a Ph.D. in Biomedical Engineering from the Georgia Institute of Technology. Her research is on data-driven solutions for simulation, healthcare, manufacturing, and sociology. Her email is \email{janani.venugopalan@thoughtworks.com}.

\end{document}